\pdfoutput=1
\documentclass[11pt]{article}

\usepackage[preprint]{acl}

\usepackage{times}
\usepackage{latexsym}

\usepackage[T1]{fontenc}
\usepackage[utf8]{inputenc}

\usepackage{microtype}

\usepackage{inconsolata}

\usepackage{url}            % simple URL typesetting
\usepackage{booktabs}       % professional-quality tables
\usepackage{amsfonts}       % blackboard math symbols
\usepackage{nicefrac}       % compact symbols for 1/2, etc.
\usepackage{microtype}      % microtypography
\usepackage{xcolor}         % colors
\usepackage{enumitem}
\usepackage{bbding}
\usepackage{array}
\newcolumntype{L}[1]{>{\raggedright\let\newline\\\arraybackslash\hspace{0pt}}m{#1}}
\newcolumntype{C}[1]{>{\centering\let\newline  \\\arraybackslash\hspace{0pt}}m{#1}}
\newcolumntype{R}[1]{>{\raggedleft\let\newline \\\arraybackslash\hspace{0pt}}m{#1}}
\usepackage{graphicx}
\usepackage{subcaption}
\usepackage{url}
\usepackage{bm}
\usepackage{amsmath,amssymb,amsfonts,amsthm}
\usepackage{caption} 
\usepackage[normalem]{ulem}
\usepackage{float} 
\usepackage{verbatim}
\usepackage{algorithm}
\usepackage{algorithmic}
\usepackage{multirow}
\usepackage{wrapfig}

\title{Case-Based Reasoning Enhances the Predictive Power of LLMs in Drug-Drug Interaction }

\author{%
	Guangyi Liu$^{1}$, Yongqi Zhang$^{2}$, Xunyuan Liu$^{1}$, Quanming Yao$^1$
	\\
	$^1$Department of Electronic Engineering, Tsinghua University \\ $^2$The Hong Kong University of Science and Technology (Guangzhou)\\
	\texttt{liugy24@mails.tsinghua.edu.cn, yzhangee@connect.ust.hk} \\
	\texttt{liuxunyu23@mails.tsinghua.edu.cn, qyaoaa@tsinghua.edu.cn} 
}

\begin{document}

\maketitle

\begin{abstract}
%\footnote{\checkmark+qm+ add abstract.}
Drug–drug interaction (DDI) prediction is critical for  treatment safety.
While large language models (LLMs) show promise in pharmaceutical tasks, 
their effectiveness in DDI prediction remains challenging. 
Inspired by 
the well-established clinical practice where physicians routinely reference similar historical cases to guide their decisions through case-based reasoning (CBR),
we propose CBR-DDI, a novel framework that distills pharmacological principles from historical cases to improve LLM reasoning for DDI tasks.
CBR-DDI constructs a knowledge repository by leveraging LLMs to extract pharmacological insights and graph neural networks (GNNs) to model drug associations.
% from KGs.
A hybrid retrieval mechanism and dual-layer knowledge-enhanced prompting allow LLMs to effectively retrieve and reuse relevant cases.
We further introduce a representative sampling strategy for dynamic case refinement.
Extensive experiments demonstrate that CBR-DDI achieves state-of-the-art performance, 
with a significant 28.7\% accuracy improvement over both popular LLMs and CBR baseline,
while maintaining high interpretability and flexibility.

\end{abstract}

\section{Introduction}

Drug-drug interaction (DDI) prediction is critical for pharmacology and healthcare, as it safeguards patients from adverse drug reactions, 
optimizes therapeutic efficacy, and reduces healthcare costs \cite{ddimagro2012epidemiology,ddiroemer2013systems,ddimarengoni2014understanding}. 
Accurately identifying DDIs is challenging due to the intricate potential relationships between drugs and the diverse mechanisms underlying the interactions
(such as the competition for drug-metabolizing enzymes) \cite{shen2024benchmarking,de2025llms}. 
These challenges become even more pronounced when predicting interactions involving new drugs, where interaction data is typically sparse or nonexistent.
\begin{figure}[t]
	\centering
	\vspace{-10px}
	\includegraphics[width=0.49\textwidth]{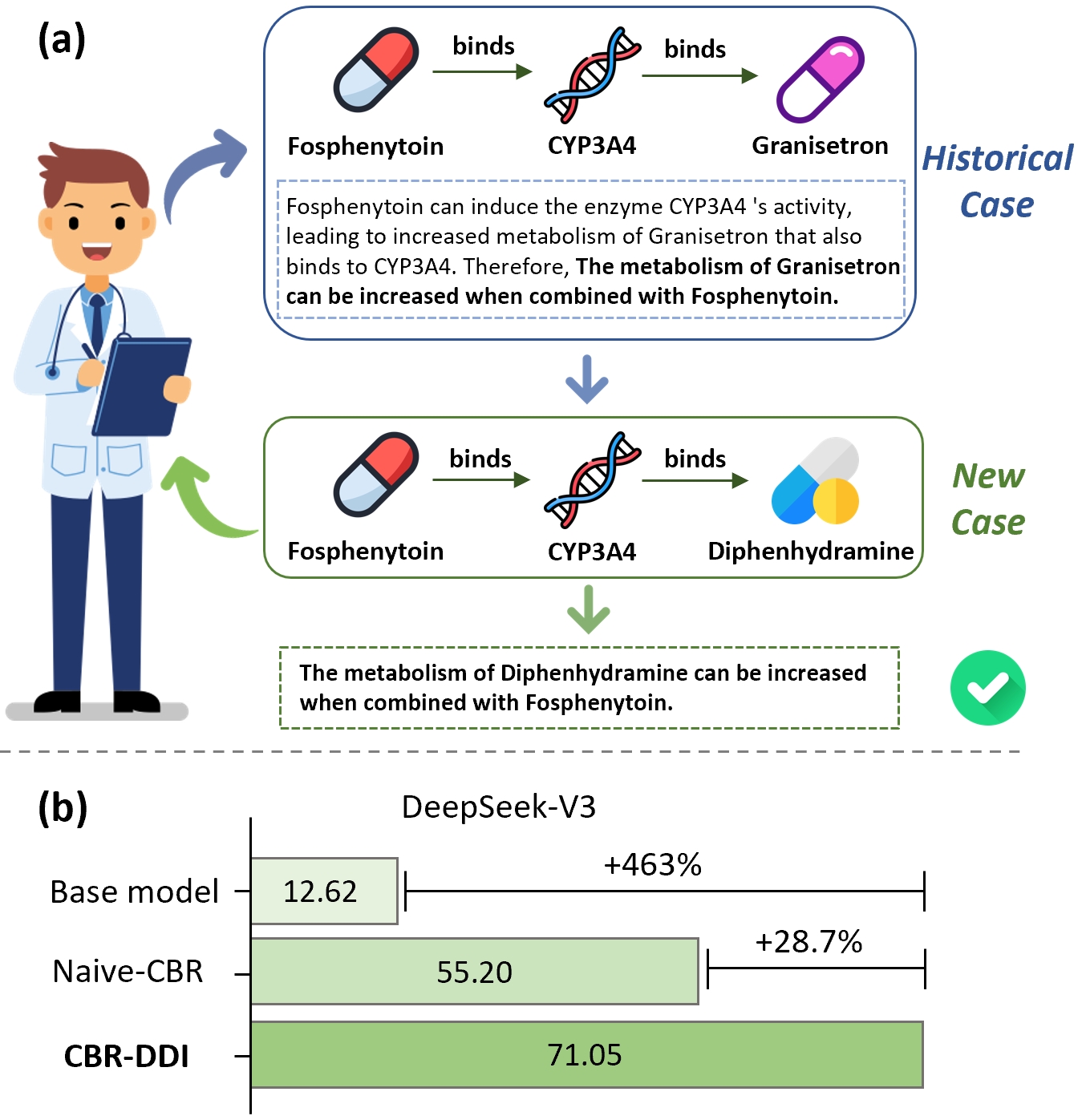}
	\vspace{-16px}
	\caption{(a).
		Illustration of using historical cases to solve new cases in DDI task.
		(b).
	Accuracy comparison on DrugBank dataset: our CBR-DDI shows significant improvement over base model and Naive-CBR.}
	\label{fig:intro}
	\vspace{-18px}
\end{figure}

Recently, large language models (LLMs) \cite{brown2020language,achiam2023gpt,grattafiori2024llama,guo2025deepseek} have demonstrated impressive capabilities across various tasks, particularly excelling at identifying hidden patterns in natural languages. 
While LLMs have shown promise in pharmaceutical applications~\cite{thirunavukarasu2023large,liang2023drugchat,inoue2024drugagent}, their effective utilization for DDI prediction remains an open research question.
%To enhance LLM's performance in this domain, 
Current approaches commonly enhance LLMs by incorporating biomedical knowledge graphs (KGs) ~\cite{ddigpt,kpath},
which provide structured knowledge about drugs.
%which describe structured relationships among drugs, genes, proteins, and diseases --- thereby providing explicit factual knowledge to support interaction prediction.
They typically employ heuristic methods to retrieve relevant drug information from KGs and feed it directly into LLMs for prediction.

However, these methods
fail to discover the underlying pharmacological mechanisms 
that explain why certain drug interactions occur~\cite{de2025llms}.
Understanding and modeling these mechanisms is essential not only for interpretability but also for generalizing predictions to new drugs~\cite{ddigpt}.
We observe that many DDI cases share common interaction mechanisms that reflect fundamental pharmacological principles among drugs. 
For instance, as illustrated in Figure~\ref{fig:intro}, a new case (drug pair \textit{Fosphenytoin}-\textit{Diphenhydramine}) and an existing case (drug pair \textit{Fosphenytoin}-\textit{Granisetron}) exhibit similar drug associations, 
enabling the transfer of known interaction mechanisms from the historical case 
to the new one. 
Yet current methods neglect these valuable inter-case relationships,
compromising the reliability and interpretability of their predictions.
This also diverges from established clinical practice \cite{althoff1998case,bichindaritz2006case}, 
where physicians routinely reference historical cases through case-based reasoning (CBR)—a cognitive process that solves new problems by adapting previously solutions to similar problems \cite{cbrwatson1994case,cbrkolodner2014case}.

Inspired by these observations, 
we propose CBR-DDI, a framework that leverages CBR 
to enhance LLMs' capabilities for DDI prediction.
Our approach constructs a structured knowledge repository 
that stores a collection of representative cases enriched with pharmacological insights.
Each case in the repository includes key associations of drug pair extracted by a GNN module from KGs, and their interaction mechanisms distilled by an LLM, providing a structured representation of pharmacological principles. 
To effectively utilize the repository, we design a hybrid retrieval strategy that identifies both semantically and structurally relevant cases, alongside a dual-layer knowledge-enhanced prompting to facilitate accurate and faithful reasoning in LLMs. 
Furthermore, to reduce storage overhead, 
we propose a sampling strategy 
that dynamically refines the repository by retaining representative cases. 
CBR-DDI achieves state-of-the-art performance across multiple benchmarks, outperforming the base LLM model by 463\% and surpassing the Naive-CBR baseline by 28.7\%. 
In addition, it offers interpretable interaction mechanisms and integrates seamlessly with off-the-shelf LLMs without requiring fine-tuning or intensive interactions.
The contributions are summarized as follows:
\begin{itemize}[leftmargin=*]
\item 
%\footnote{\checkmark+qm+ check footnote~\ref{ft:1}.}
Inspired by the success of CBR in clinical practice,
we propose CBR-DDI, a new framework that distills pharmacological principles from historical cases to enhance LLM’s reasoning  for DDI tasks.

\item
We propose to construct a knowledge repository, through a collaboration between LLMs for distilling pharmacological insights and GNNs for extracting drug associations from biomedical knowledge graphs.

\item
For the deployment of the knowledge repository, we design a hybrid retrieval mechanism to identify relevant cases, a dual-layer knowledge-enhanced prompting to guide LLMs in case reuse, and a representative sampling strategy for repository refinement.

\item
Extensive experiments on DDI demonstrate CBR-DDI achieves state-of-the-art performance while maintaining high interpretability and flexibility.	
\end{itemize}

\section{Related Work}

%\subsection{Drug-Drug Interaction Prediction}
\textbf{Drug-Drug Interaction Prediction.}
The task of DDI prediction identifies potential adverse interactions or synergistic effects between co-administered medications \cite{ddimagro2012epidemiology,ddiroemer2013systems}. 
Measuring DDIs in clinical experiments is time-consuming and costly, driving the adoption of machine learning approaches \cite{shen2024benchmarking,ddireviewluo2024drug}.
\textit{Feature-based methods} leverage shallow models to classify DDI types using drug pair features (e.g., fingerprints) \cite{finger1rogers2010extended,finger2ryu2018deep}. 
\textit{Graph-based methods} model the drug interaction data as a graph.
Simple approaches employ embedding techniques \cite{complextrouillon2017knowledge,yao2022effective} to learn drug representations.
More advanced methods enhance prediction by incorporating biomedical KGs \cite{kg1himmelstein2015heterogeneous,kg2chandak2023building}, 
which represent relationships between biomedical concepts (e.g., drugs, genes, and diseases) in a multi-relational structure.
To capture structural patterns in the graph, various deep models have been proposed, such as graph neural networks (GNNs) \cite{Decagonzitnik2018modeling,kgnnlin2020kgnn,sumgnn2021,zhang2023emerging} and graph transformers \cite{tigersu2024dual}.
\textit{Language model (LM)-based methods} \cite{textddizhu2024learning} leverage drug descriptions to train models (e.g., RoBERTa \cite{liu2019roberta}) for prediction.
Notably, another category of methods \cite{mol1chen2021muffin,mol3zhong2024learning,sun2025exddi} uses drug molecular structures as input, whereas our approach does not, making these methods orthogonal to ours.

\begin{figure*}[ht]
	\centering
	\includegraphics[width=0.85\textwidth]{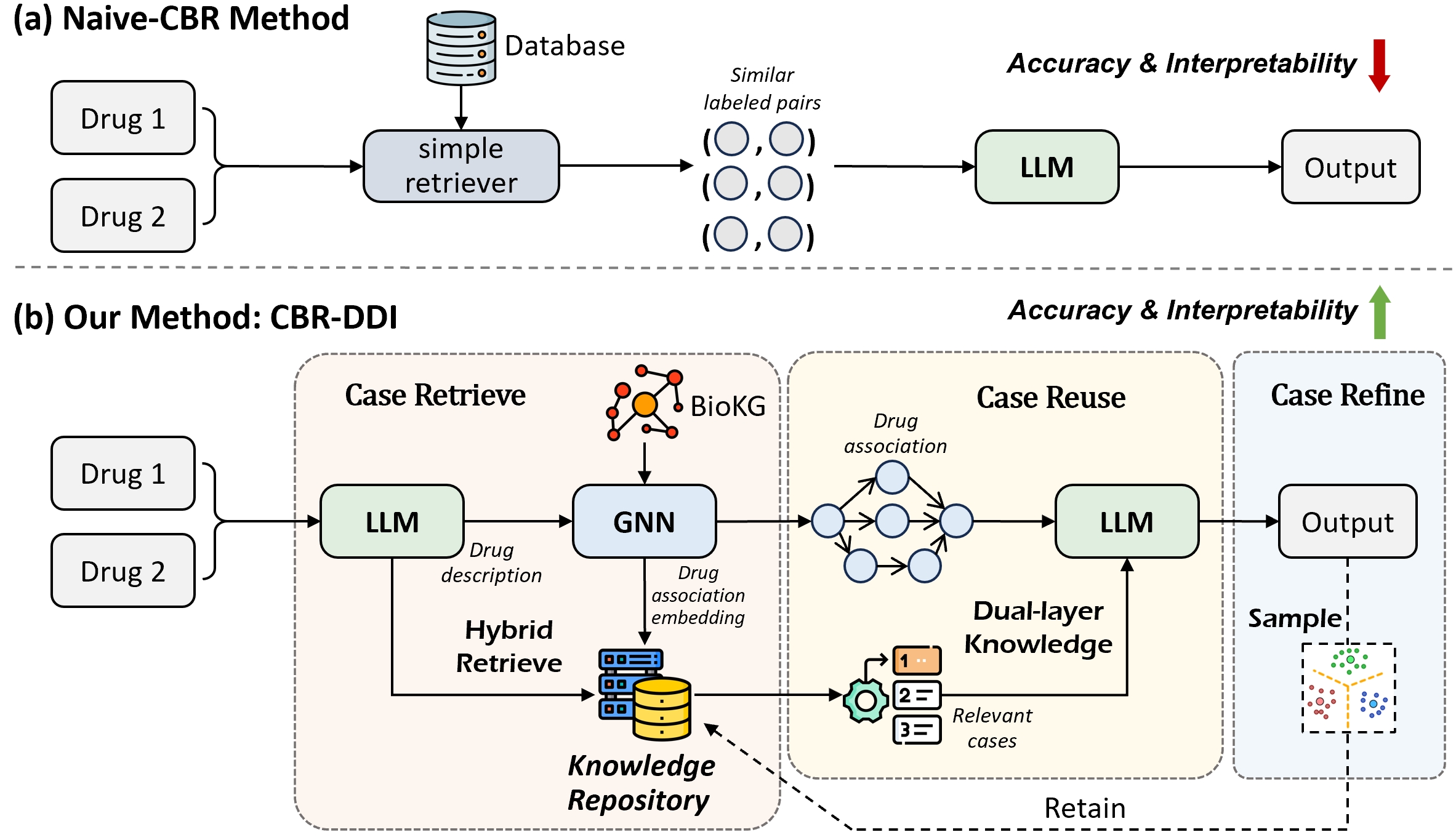}
	\vspace{-4px}
	\caption{Comparison between Naive-CBR method and our method CBR-DDI. CBR-DDI constructs a knowledge repository storing cases with rich pharmacological insights, and enhances LLM predictions via LLM-GNN collaborative case retrieval, dual-layer knowledge-enhanced reuse, and representative sampling-based dynamic refinement.	}
	\label{fig:overview}
	\vspace{-14px}
\end{figure*}

Recently, LLMs are increasingly utilized in biomedical applications, including drug discovery~\cite{chaves2024tx}, repurposing~\cite{inoue2024drugagent}, and molecular understanding~\cite{liang2023drugchat}. Their pre-training on vast biomedical literature enables them to leverage implicit knowledge about drug interactions~\cite{sun2025exddi,de2025llms}.
However, complex drug associations, diverse interaction mechanisms, and multiple interaction types pose significant challenges for LLMs in DDI prediction. 
Recent approaches heuristically retrieve drug information (e.g., paths between drugs~\cite{kpath}, one-hop neighbors~\cite{ddigpt}) from KGs 
and feed it directly into LLMs. 
However, they fail to discover the underlying pharmacological mechanisms, reducing the reliability and generalization to new drug prediction.

\noindent
\textbf{Retrieval-Augmented Generation.}
Retrieval-Augmented Generation (RAG)~\cite{gao2023retrieval,huang2024survey,yang2024buffer} is a framework that enhances the generative capabilities of LLMs by retrieving relevant knowledge from an external knowledge source.
%, offering a promising solution to mitigate the  hallucination of LLMs.
Recent advancements have explored to retrieve from KGs to enhance LLMs' reasoning~\cite{pan2024roadmap,agrawal2023can}.
These methods primarily extracting question-relevant reasoning paths from KGs for LLMs~\cite{luo2023reasoning,sun2023tog}. 
However, in DDI tasks, explicit questions are absent, and the diverse relational paths between drugs do not directly reveal their interaction type, making these methods challenging to adapt effectively.

\noindent
\textbf{Case-Based Reasoning (CBR).}
CBR is a problem-solving paradigm that addresses new problems by adapting solutions from previously resolved cases~\cite{cbrslade1991case,cbrwatson1994case,cbrkolodner2014case}. 
Typical CBR process involves retrieving similar past problems, reusing their solutions, evaluating the effectiveness, revising the solution, and retaining successful solutions~\cite{cbrwatson1994case}. 
Historically, CBR has been widely applied across various domains, such as medical diagnosis~\cite{koton1988using},
and industrial problem-solving~\cite{hennessy1992applying}. 
%and decision support systems~\cite{goodman1989cbr}. 
Recently, there has been increasing interest in integrating CBR with LLMs~\cite{wilkerson2024implementing,yang2024casegpt,guo2024ds}.
% enabling advancements in tasks such as legal question answering~\cite{cbrwiratunga2024cbr} and automated data science~\cite{guo2024ds}.
However, applying CBR to the DDI task is non-trivial, 
as it requires carefully designed case retrieval strategies, and existing datasets typically contain only interaction labels without in-depth pharmacological insights as solutions that can be transferred to new cases.

\section{Proposed Method}

\subsection{Overall Framework}

In DDI prediction task, we have a set of drugs $\mathcal{V_D}$ and interaction relations $\mathcal{R_D}$
among them. 
Given a query drug pair $(u, v)$, the goal of DDI prediction is to determine their interaction type $r \in \mathcal{R_D}$.
We formulate it as a reasoning task for LLMs to select the most likely interaction type $r$ from the relation set $\mathcal{R_D}$.
Additionally, we utilize a biomedical KG
to capture the associations of drugs.
%$\mathcal{G} = \{(h, r, t)\}$, where entities $h$ and $t$ include drugs, diseases, genes, and other biomedical concepts, and relations $r$ comprise both drug interaction types $\mathcal{R_D}$ and broader biomedical relations $\mathcal{R_B}$. 

While the diversity of interaction mechanisms presents a significant challenge for DDI prediction, 
different cases may share interaction patterns, 
reflecting universal pharmacological principles
\cite{tummino2021drug,roberti2021pharmacology}. 
Inspired by the proven success of CBR in clinical practice, we propose CBR-DDI, a framework that 
distills pharmacological principles from historical cases to enhance LLM’s reasoning.
In contrast to naive CBR applications~\cite{brown2020language} that rely on simple retrieval methods (e.g., fingerprint-based matching~\cite{finger1rogers2010extended}) and offer only interaction labels as solutions,
CBR-DDI constructs a knowledge repository that integrates rich pharmacological insights, and strengthens LLMs through comprehensive case retrieval, knowledge-enhanced reuse, and dynamic refinement of resolved cases.

As illustrated in 
Figure~\ref{fig:overview}, the framework operates in three stages: 
(1) case retrieval via LLM-GNN collaboration, 
(2) case reuse via dual-layer knowledge guided 
reasoning, 
and (3) case refinement via representative sampling.
Given the names of a drug pair, 
we first leverage the LLM to generate concise drug descriptions, 
which are used both to perform semantic-level retrieval and to augment a GNN module that encodes the subgraph of the drug pair in the KG.
This enables a hybrid retrieval mechanism that identifies both semantically and structurally relevant cases 
from the knowledge repository. 
Then, the retrieved cases are integrated into a dual-layer knowledge-enhanced prompt, which combines key drug associations extracted by the GNN module with historically similar interaction mechanisms, guiding the LLM to generate accurate and explainable prediction.
Finally, we design a sampling strategy to refine the repository by grouping similar cases and retaining representative ones, reducing redundancy and improving adaptability to new discoveries.

\subsection{Knowledge Repository}
\label{ssec:repo}

\begin{figure}[t]
	\centering
	\vspace{-2px}
	\includegraphics[width=0.48\textwidth]{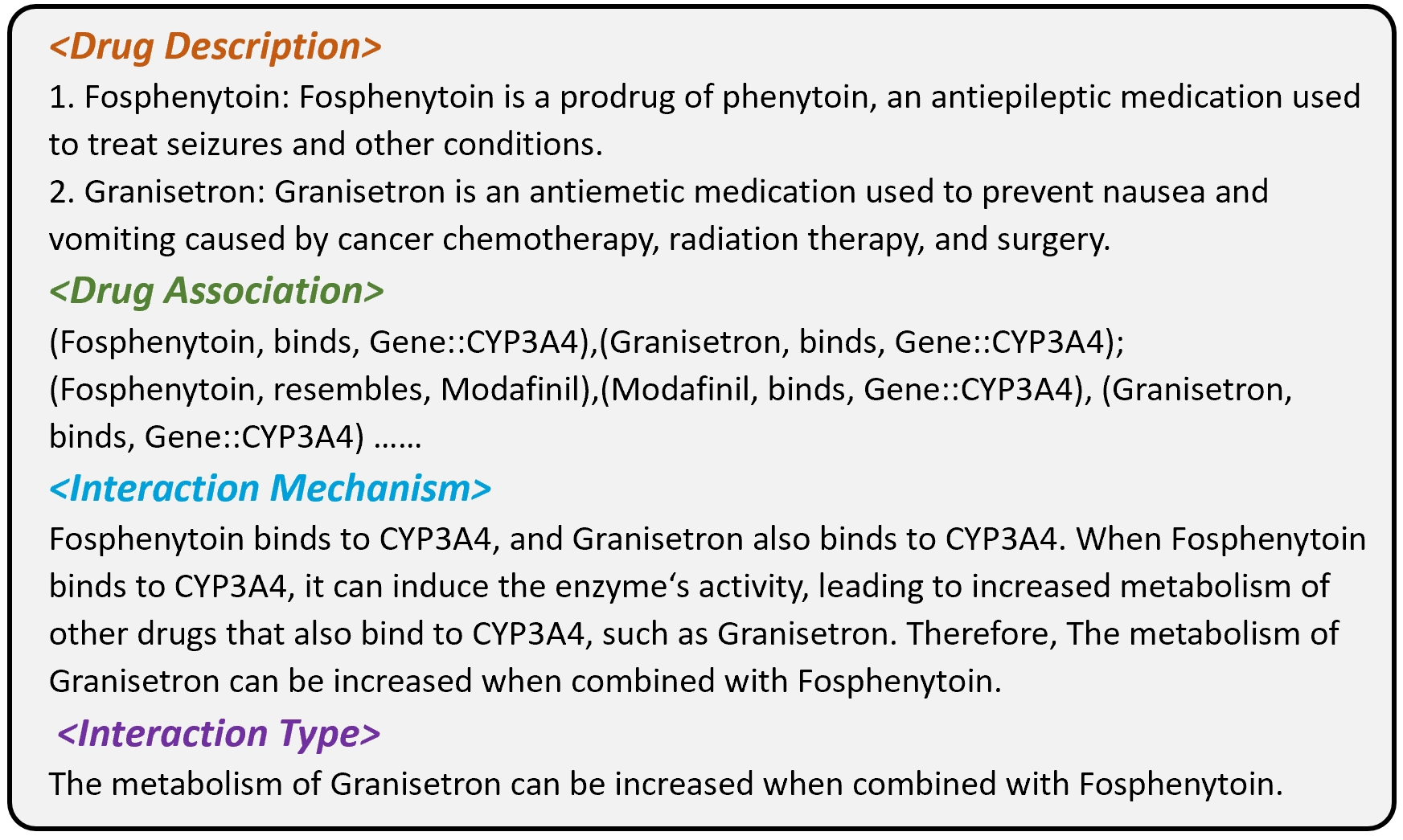}
	\vspace{-22px}
	\caption{Example from the knowledge repository.}
	\label{fig:case0}
	\vspace{-18px}
\end{figure}

To effectively leverage the historical drug interaction cases and discover important pharmacological principles, 
we propose to construct a lightweight knowledge repository that stores 
a collection of representative cases enriched with pharmacological insights.
This design is inspired by the case-based reasoning paradigm widely adopted in clinical decision support systems~\cite{althoff1998case,bichindaritz2006case}, where past cases are enriched and reused to guide new decisions.
The repository is designed to capture both factual information of drugs and generalizable pharmacological patterns, thereby enabling accurate retrieval of relevant cases and facilitating analogical reasoning in predicting new drug interactions.
Specifically,
as shown in Figure~\ref{fig:case0},
each case $C$ involving a drug pair $(u,v)$ in the repository is a structured representation of DDIs, consisting of four key components:
\begin{itemize}[leftmargin=*]
	\item drug description $D_c=(D_u, D_v)$: functional descriptions of the drugs generated by LLM (detailed in Section~\ref{ssec:retrieve}).;	
	\item drug association $H_c$: structured knowledge extracted from the KG using the GNN module, representing the relationships between drugs, with representation $\bm h_c$ (detailed in Section~\ref{ssec:reuse});
	\item interaction mechanism $M_c$: pharmacological insights that explain why the drugs interact, distilled from domain knowledge and historical cases by LLM (detailed in Section~\ref{ssec:reuse});
	\item interaction type $T_c$: the label of interaction;
\end{itemize}
Among these, drug descriptions and associations provide factual grounding for retrieval, while the interaction mechanism is the core of each case, as it explains the underlying reason for the interaction, providing key pharmacological principles that can be transferred to the prediction of new drug pairs. 
%More cases are provided in Appendix~\ref{appendix:case}.

\subsection{Reasoning Steps}
\label{ssec:step}

\subsubsection{Case Retrieval via LLM-GNN Collaboration}
\label{ssec:retrieve}

Effective case retrieval is crucial for CBR, as the relevance and quality of retrieved cases directly impact the accuracy and interpretability of predictions. Considering the diverse functions of drugs and their varying associations, we propose a hybrid retrieval mechanism that combines the natural language processing capabilities of LLMs with the structured learning abilities of GNNs,
enabling retrieval of semantically and structurally similar cases. 
%from knowledge repository.
%ensuring that retrieved cases share similar pharmacological properties and  biomedical associations.

To retrieve relevant historical cases $C$'s for a given drug pair $p = (u, v)$, we compute a retrieval score based on a weighted combination of semantic similarity and structural similarity: 
\begin{equation} 
 	\begin{split}
	s(p,c) = {} & \lambda \cdot \text{SemanticSim}(p,c) \\ 
	& + (1-\lambda) \cdot \text{StructSim}(p, c),
	\end{split}
	\label{eq:retrieve} 
\end{equation} 
where $\lambda \in [0,1]$ is a hyperparameter that balances the contribution of the two components. 
The two similarity are defined as follows: 
\begin{itemize}[leftmargin=*] 
	\item 
	$\text{SemanticSim}(p,c) = \text{Sim}(f(D_p),f(D_c))$:
	We prompt an LLM (i.e., Llama3.1-8B-Instruct~\cite{grattafiori2024llama}) to generate concise functional descriptions $D_u$ and $D_v$ for drugs $u$ and $v$, denoted as $D_p = (D_u, D_v) = \text{LLM}_{\text{des}}(u,v)$. The function $f(\cdot)$ denotes a text embedding model \cite{liu2019roberta}. We then compute the cosine similarity between the embeddings of $D_p$ and the stored case description $D_c$, capturing the semantic closeness of drug functionality and pharmacological properties.
	
	\item 
	$\text{StructSim}(p,c) = \text{Sim}(\bm h_p, \bm h_c)$:
	We employ a subgraph-based GNN module with attention mechanism (i.e., EmerGNN~\cite{zhang2023emerging}) to encode the subgraph connecting the drug pair in KG, with the embeddings of LLM-generated drug descriptions as node features, obtaining the subgraph representation:
	$\bm h_p = \text{GNN}(f(D_u),f(D_v))$.
	Cosine similarity is then computed between $\bm h_p$ and the stored case representations $\bm h_c$, 
	reflecting the structural similarity in the association patterns between drug pairs.
\end{itemize}

We rank all cases in the repository based on $s(p, c)$ and select the top-$K$ most relevant ones for subsequent reasoning.
By integrating semantic drug descriptions with graph-structured relational knowledge,
this hybrid approach enables a comprehensive case retrieval process, capturing pharmacologically similar drug pairs while preserving structural association relevance.

\subsubsection{Case Reuse via Dual-layer Knowledge Guided Reasoning}
\label{ssec:reuse}

Although relevant cases reflect potential interaction mechanisms, they do not provide sufficient factual information for the given drug pair. 
To address this, we design a dual-layer knowledge-enhanced prompt that integrates both external factual knowledge (i.e., drug associations) and internal regularity knowledge (i.e., historical interaction mechanisms) to guide the LLM's reasoning process. 

Specifically, the prompt comprises the key drug associations of given pair extracted by the attention-based GNN module, and relevant interaction mechanisms contained in historical similar cases.
The LLM is then prompted to synthesize these two complementary sources of knowledge, generating the interaction mechanism $M_p$ and type $T_p$.
The prediction process is formalized as:
\begin{equation}
M_p, T_p = \text{LLM}_{\text{pre}}(TD, \{C_i\}_{i=1}^K, H_p, A_p),
	\label{eq:predict} 
\end{equation}
where $TD$ is the task description, $\{C_i\}_{i=1}^K$ are the top-$K$ retrieved cases, $H_p$ denotes the extracted drug association facts, and $A_p$ is the filtered candidate interaction types.
We detail the two types of knowledge as follows: 
\begin{itemize}[leftmargin=*] 
	\item External factual knowledge (i.e., drug associations $H_p$):
	To capture essential associations between drugs, we employ the attention-based GNN module to extract high-quality relational paths that connect them.
	Unlike prior work~\cite{kpath} that retrieves triplets heuristically, we scores triplets along the paths by attention weights during GNN propagation.
	We then select the top-$P$ paths with the highest average attention as $H_p$, which are incorporated into the prompt as structured, high-quality factual evidence
	(e.g., {\small \textit{Fosphenytoin $\xrightarrow{binds}$ CYP3A4 $\xrightarrow{binds}$ Diphenhydramine}}).
	
	\item Internal regularity knowledge (i.e., interaction mechanisms within historical cases $\{C_i\}_{i=1}^K$ ):  
	The retrieved cases (in Section~\ref{ssec:retrieve}) contain interaction mechanisms $M_{c_i}$ that reflect generalized pharmacological patterns observed in similar drug pairs.  
	They can guide the LLM to perform analogical reasoning, drawing parallels between the current drug pair and previously known interaction regularity.  
\end{itemize}

By structuring the prompt in this manner, we enhance the interpretability and reliability of LLM-generated predictions, as the historical cases offer relevant pharmacological principles, while the factual drug associations provide the evidence base. 
Furthermore,
to reduce the complexity introduced by numerous interaction types, we pre-filter candidate answers $A_p$ based on the scores of GNN module, retaining only top-$N$ candidates.
This focuses the LLM’s attention on the most plausible options and reduces noise from irrelevant candidates.

%Furthermore, filtering candidate answers refines the model’s decision-making process, improving both efficiency and prediction accuracy.

\subsubsection{Case Refinement via Representative Sampling}

To ensure both the quality and size control of our knowledge repository, we propose a dynamic refinement strategy that updates cases in the knowledge repository.
Specifically,
for each LLM-generated prediction, we verify its correctness against ground truth label (e.g., from training data or expert feedback), and
prompt revisions for errors based on the correct label. 
Furthermore, 
to control the growth of the repository while preserving its expressive power,
we group semantically similar cases within each DDI category using the text embeddings of their interaction mechanisms $M_c$. 
Our case-based design allows for simple yet effective clustering methods to retain only the most representative cases—filtering out redundancy while preserving diversity in pharmacological scenarios
(details are shown in Appendix~\ref{appendix:detail of repo}).
This approach keeps the repository compact and efficient while allowing for new discoveries.
%, effectively balancing quality control with growing demands. 

\subsection{Comparison with Existing Works}
\label{ssec:comparison}

\begin{table}[t]
	\centering
	\vspace{-4px}
	\footnotesize
%	\scalebox{0.75}{
	\begin{tabular}{c|cccc}
		\toprule
		Methods       & WFT  & ITP & DAA & IMA \\ \midrule
		TextDDI      & $\times$  & $\times$         & $\times$                      & $\times$                           \\
		DDI-GPT    & $\times$            &  \checkmark       & $\times$                      & $\times$                           \\
		Naive-CBR  & \checkmark          &  $\times$         & $\times$                      & $\times$                           \\
		K-Paths              & \checkmark                           & \checkmark       & \checkmark                    & $\times$                           \\ \midrule
		CBR-DDI       & \checkmark           & \checkmark       & \checkmark                    & \checkmark                         \\ \bottomrule
	\end{tabular}
%}
	\vspace{-8px}
	\caption{Comparison of different methods using LMs. WFT: Without Fine-Tuning; ITP: Interpretability; DAA: Drug Association Augmentation; IMA: Interaction Mechanism Augmentation.}
	\label{tab:comparison}
	\vspace{-18px}
\end{table}

As shown in Table~\ref{tab:comparison},
TextDDI~\cite{textddizhu2024learning} and DDI-GPT~\cite{ddigpt} rely on fine-tuning small language models (e.g., RoBERTa~\cite{liu2019roberta}) as classifiers, which limits their compatibility with off-the-shelf LLMs.
Specifically, TextDDI 
%ignores drug association and interaction mechanism knowledge, 
relys solely on individual drug descriptions.
DDI-GPT retrieves one-hop neighbors from KGs for binary classification and applies an attention mechanism for limited interpretability.
Naive-CBR method~\cite{brown2020language} retrieves structurally similar drug pairs based on fingerprint features, providing only case labels for LLMs without deeper pharmacological insight.
K-Paths~\cite{kpath} uses heuristic methods to extract diverse paths between drugs and directly feeds them into LLMs.
%However, these approaches neglect the importance of pharmacological mechanisms derived from historical cases. Without such regularity knowledge, LLMs may struggle to perform accurate reasoning based solely on structural associations.
In contrast, CBR-DDI uniquely integrates both drug association knowledge and interaction mechanism knowledge to augment LLM reasoning,
enabling accurate and interpretable prediction, while offering plug-and-play flexibility across LLMs without requiring fine-tuning.

\begin{table*}[t]
	\centering
	%	\vspace{-8px}
	%	\renewcommand\arraystretch{0.98}
	\setlength{\tabcolsep}{5pt}
	\vspace{-8px}
	\footnotesize
	\begin{tabular}{c|l|cccc|cccc|c}
		\toprule
		    \multirow{3}{*}{Type}     & \multirow{3}{*}{Method} &                  \multicolumn{4}{c|} {DrugBank}                   &                    \multicolumn{4}{c|}{ {TWOSIDES}}                     & \multirow{3}{*}{$\Delta_{avg}$} \\
		      &                         &     \multicolumn{2}{c}{S1}      &     \multicolumn{2}{c|}{S2}     &      \multicolumn{2}{c}{ {S1}}      &     \multicolumn{2}{c|}{{S2}}     &                                 \\
		 &           &     {Acc}      &       F1       &     {Acc}      &       F1       &     {Recall}     &      {NDCG}      &    {Recall}     &     {NDCG}      &                                 \\ 
		 \midrule
		 Feature-based         & MLP                     &     57.77      &     42.53      &     39.85      &     20.15      &      12.70       &      14.88       &      3.60       &      5.95       &         6.42 $\uparrow$         \\ \midrule
		\multirow{6}{*}{Graph-based}  & ComplEx                 &      4.02      &      1.74      &      4.32      &      1.77      &       2.30       &       3.61       &      1.62       &      1.81       &        32.06 $\uparrow$         \\
		& MSTE                    &     54.66      &     40.57      &     32.88      &      4.93      &       5.12       &       7.37       &      2.78       &      3.12       &        11.02 $\uparrow$         \\
		& Decagon                 &     32.41      &     28.56      &     22.47      &      6.12      &       4.48       &       6.36       &      2.38       &      3.61       &        19.54 $\uparrow$         \\
		& SumGNN                  &     57.04      &     54.77      &     25.28      &     17.85      &       4.08       &       5.24       &      2.11       &      3.48       &        13.03 $\uparrow$         \\
		& {EmerGNN}               &    {68.10}     &     65.78      &    {44.84}     &     34.22      &     {13.79}      &     {16.06}      &     {3.01}      &     {4.93}      &         2.45 $\uparrow$         \\
		& TIGER                   &     60.11      &     57.21      &     33.46      &     19.78      &      11.72       &      14.33       &      2.69       &      3.90       &         7.81 $\uparrow$         \\ \midrule
		LM-based            & TextDDI                 &     66.75      &     66.53      &     44.23      &     32.79      &       9.88       &      13.24       &      4.16       &      6.04       &         3.35 $\uparrow$         \\ \midrule
		\multirow{4}{*}{Llama3.1-8B}  & Base                    &      8.71      &      4.10      &      7.30      &      3.94      &       0.04       &       0.06       &      0.02       &      0.03       &        28.92 $\uparrow$         \\
		& Naive-CBR               &     47.88      &     42.38      &     15.02      &      8.70      &       3.60       &       4.47       &      0.27       &      0.50       &        16.24 $\uparrow$         \\
		& K-Paths                 &     17.62      &      9.06      &     12.29      &      7.34      &       0.25       &       0.38       &      0.07       &      0.08       &        25.38 $\uparrow$         \\
		& CBR-DDI                 & \textbf{68.52} & \textbf{61.57} & \textbf{44.94} & \textbf{32.43} & \textbf{{13.89}} & \textbf{{15.45}} & \textbf{{4.38}} & \textbf{{7.04}} &                -                \\ \midrule
		\multirow{4}{*}{Llama3.1-70B} & Base                    &      8.93      &      4.37      &      8.02      &      4.12      &       0.05       &       0.06       &      0.03       &      0.03       &        30.21 $\uparrow$         \\
		& Naive-CBR               &     48.09      &     50.62      &     21.22      &     13.04      &       4.54       &       5.46       &      0.68       &      0.84       &        15.84 $\uparrow$         \\
		& K-Paths                 &     31.35      &     16.43      &     31.12      &     14.87      &       2.09       &       3.18       &      1.01       &      1.42       &        18.08 $\uparrow$         \\
		& CBR-DDI                 & \textbf{71.36} & \textbf{70.85} & \textbf{47.43} & \textbf{36.88} &  \textbf{14.40}  &  \textbf{16.97}  &  \textbf{4.68}  &  \textbf{7.32}  &                -                \\ \midrule
		& Base                    &     12.62      &      9.61      &     12.12      &      6.78      &       0.03       &       0.04       &      0.03       &      0.05       &        28.82 $\uparrow$         \\
		{DeepSeek-V3}         & Naive-CBR               &     55.20      &     47.24      &     22.26      &     15.46      &       3.18       &       4.22       &      0.32       &      0.47       &        14.78 $\uparrow$         \\
		-671B             & K-Paths                 &     34.52      &     18.17      &     32.33      &     15.41      &       1.73       &       2.21       &      1.19       &      1.66       &        17.58 $\uparrow$         \\
		& CBR-DDI                 & \textbf{71.05} & \textbf{74.38} & \textbf{49.45} & \textbf{40.69} &  \textbf{14.85}  &  \textbf{16.56}  &  \textbf{4.73}  &  \textbf{6.60}  &                -                \\ \bottomrule
	\end{tabular}
	\vspace{-6px}
	\caption{\label{tab:result}
	Performance comparison of different methods for DDI. $\Delta_{avg}$ denotes the average improvement in accuracy and recall (in percent) on two datasets.}
	\vspace{-10px}
\end{table*}

\section{Experiment}

\subsection{Experimental Setup}

\noindent
\textbf{Datasets.}
We conduct experiments on two widely used DDI datasets: (1) DrugBank~\cite{drugbankwishart2018drugbank}, a multi-class  dataset that contains 86 types interactions between drugs. 
(2) TWOSIDES~\cite{twotatonetti2012data}, 
a multi-label dataset that records 200 side effects between drugs.

\noindent
\textbf{Experimental Settings.}
Following \cite{zhang2023emerging,kpath,dewulf2021cold}, 
we evaluate our model on two challenging settings: S1 and S2.
For S1 setting, the task is to predict the interaction type between an emerging drug—one that has no interaction records in the training set—and an existing drug.
For S2 setting, the goal is to predict the interaction type between two emerging drugs.
We also provide experimental results for S0 setting in Appendix~\ref{appendix:S0}.

\noindent
\textbf{Evaluation Metrics.}
For the DrugBank dataset, where each drug pair corresponds to a single interaction type, we adopt Accuracy and F1 Score as evaluation metrics.
For the TWOSIDES dataset, where a drug pair may involve multiple interaction types, we treat it as a recommendation task and use Recall@5 and NDCG@5 as the evaluation metrics.

\noindent
\textbf{Experiment Details.}
We follow the settings of \cite{zhang2023emerging} to train the GNN module
and use HetioNet~\cite{kg1himmelstein2015heterogeneous} as the external KG.
Considering the plug-and-play convenience of CBR-DDI, we use
three LLMs in experiments: Llama3.1-8B-Instruct~\cite{grattafiori2024llama}, Llama3.1-70B-Instruct~\cite{grattafiori2024llama},  
and DeepSeek-V3~\cite{liu2024deepseek}.
We typically set number of reference cases $K$ as 5, the number of paths in drug associations $P$ as 5, and vary the number of candidate answers among \{3,5,10\}.
%The training of GNN module and the inference of Llama3.1-8B are on an RTX 3090-24GB GPU, while the inference for Llama3.1-70B runs on two A100-80GB GPUs.
%DeepSeek is accessed via API calls.
Other details are shown in Appendix~\ref{appendx:exp}.

\noindent
\textbf{Baseline Methods.}
We consider the following baseline methods for comparison: 
(1) traditional methods without using LLMs: 
MLP~\cite{mlp1998artificial},
ComplEx~\cite{complextrouillon2017knowledge},
MSTE~\cite{yao2022effective},
Decagon~\cite{Decagonzitnik2018modeling},
SumGNN~\cite{sumgnn2021}, EmerGNN~\cite{zhang2023emerging},
TIGER~\cite{tigersu2024dual},
TextDDI~\cite{textddizhu2024learning};
(2) LLM-based methods:
Base model,
Naive-CBR (retrieve 10 similar labeled cases based on fingerprint similarity as few-shot prompting~\cite{brown2020language}),
K-Paths~\cite{kpath}.

\subsection{Performance Comparison}

As shown in Table~\ref{tab:result}, among LLM-based baselines, Naive-CBR achieves notable performance improvements, highlighting the importance of historical cases in prediction. 
By providing similar drug pairs with their interaction labels, it demonstrates that past interaction patterns offer valuable knowledge for guiding LLM predictions. 
However, Naive-CBR relies on untrained and simple feature similarity metrics, which fail to capture complex relationships between cases or provide in-depth pharmacological insights. Consequently, it can not outperform other advanced deep learning approaches that are specifically trained for DDI.
In contrast, our proposed method, CBR-DDI, significantly outperforms all baseline methods across multiple benchmarks, especially when paired with powerful LLMs like Llama3.1-70B or DeepSeek. 
Even with smaller models such as Llama3.1-8B, our method achieves superior results over state-of-the-art methods. 
Compared to heuristic-based approaches like K-Paths, which may introduce irrelevant or redundant information,
CBR-DDI effectively leverages historical cases to extract valuable pharmacological insights, and enhances LLM outputs by integrating both factual drug association knowledge and regular interaction mechanism knowledge, thereby achieving more accurate and reliable predictions.
These results demonstrate that CBR-DDI is the first work to effectively unlock the potential of LLMs for DDI prediction.

\subsection{Ablation Study}

\begin{table}[t]
\centering
\setlength{\tabcolsep}{2.6pt}
\footnotesize
\begin{tabular}{c|cccc|cccc}
	\toprule
	   \multirow{3}{*}{\shortstack{CBR\\-DDI}}&   \multicolumn{4}{c|}{DrugBank}           &   \multicolumn{4}{c}{TWOSIDES}           \\
	     & \multicolumn{2}{c}{S1} & \multicolumn{2}{c|}{S2} & \multicolumn{2}{c}{S1} & \multicolumn{2}{c}{S2} \\
	   & {Acc} &      {F1}      & {Acc}  &      {F1}      & {Rec} &   {NDCG}    & {Rec} &   {NDCG}    \\ \midrule
	  full   & 71.4  &     {70.9}     & {47.4} &     {36.9}     &  {14.4}  &   {17.0}    &  {4.7}   &    {7.3}    \\
	w.o.case & 68.3  &      68.4      &  46.0  &      33.5      &   13.9  &    15.1     &   3.4   &     5.2     \\
	w.o.asso & 69.4  &      68.9      &  46.5  &      34.2      &   14.1   &    16.4     &   4.4   &     7.0     \\ \bottomrule
\end{tabular}
	\vspace{-8px}
	\caption{\label{tab:ablation}
	Comparison of different variants of CBR-DDI-Llama3.1-70B.}
\vspace{-10px}
\end{table}

\begin{table}[t]
	\centering
%	\vspace{-6px}
	%	\renewcommand\arraystretch{0.98}
	\setlength{\tabcolsep}{1.2pt}
	\footnotesize
	\begin{tabular}{l|cccc|cccc}
		\toprule
		\multirow{3}{*}{CBR-DDI} & \multicolumn{4}{c|} {DrugBank} & \multicolumn{4}{c}{ {TWOSIDES}}\\                                                 
		& \multicolumn{2}{c}{S1}  & \multicolumn{2}{c|}{S2} & \multicolumn{2}{c}{{S1}} & \multicolumn{2}{c}{{S2}} \\
		& {Acc}  & \#Case   & {Acc} & \#Case & {Rec} & \#Case & {Rec} & \#Case    \\
		\midrule
		w.o.samp  & 71.36 &35255 & 47.38 & 3056 & 14.32 & 4684 & 4.68  & 808\\
		w.samp  & 71.05  & \textbf{2139} & 47.43 & \textbf{398} & 14.40 & \textbf{1639} &4.48 & \textbf{504}\\
		\bottomrule
	\end{tabular}
	\vspace{-8px}
	\caption{\label{tab:cluster}
		Influence of representative sampling strategy.}
	\vspace{-16px}
\end{table}

\begin{figure}[t]
	\centering
	\vspace{-8px}
	\begin{subfigure}[t]{0.235\textwidth}  
		\centering
		\includegraphics[width=\linewidth]{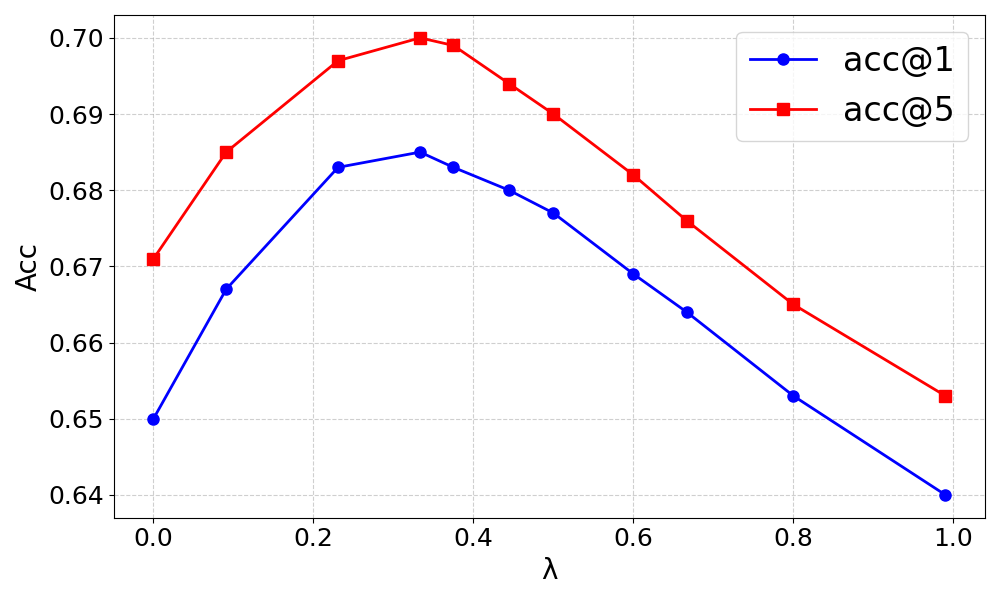}
		\caption{Acc vs $\lambda$ on DrugBank-S1}
		\label{fig:lambda_s1}
		\vspace{-5pt}
	\end{subfigure}
%	\hfill
	\begin{subfigure}[t]{0.235\textwidth}  
		\centering
		\includegraphics[width=\linewidth]{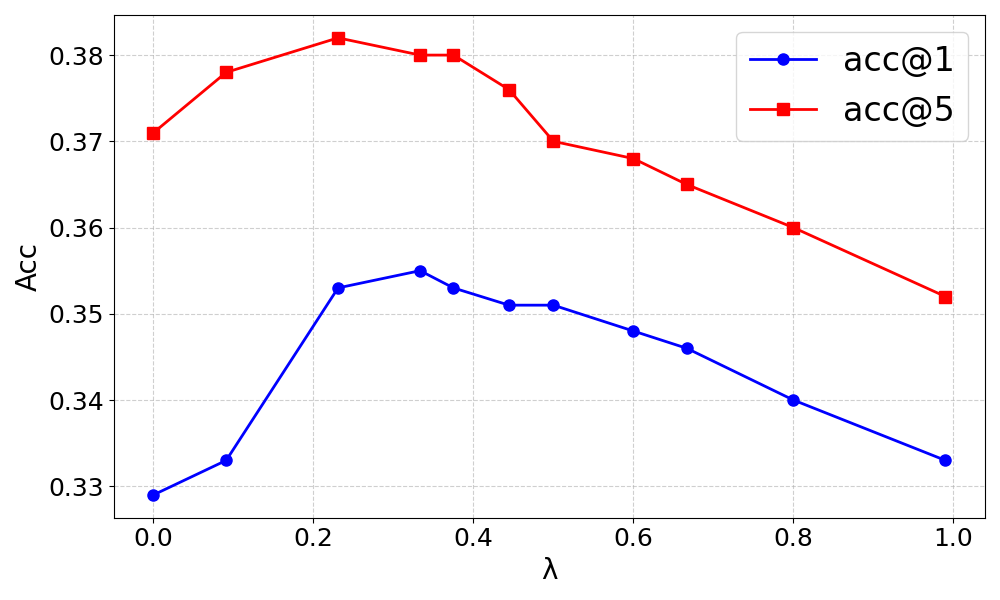}
		\caption{Acc vs $\lambda$ on DrugBank-S2}
		\label{fig:lambda_s2}
		\vspace{-5pt}
	\end{subfigure}
	\caption{Impact of hybrid retriever's hyperparameter.}
	\label{fig:hybrid}
	\vspace{-16px}
\end{figure}

\subsubsection{Influence of Dual-Layer Knowledge Augmentation}
To validate the necessity of both factual knowledge (i.e., drug associations) and regularity knowledge (i.e., interaction mechanisms derived from cases), we conduct ablation studies under three configurations: 
(i) the full prompt, 
(ii) factual-only (w.o. case), 
and (iii) regularity-only (w.o. asso).
As shown in Table~\ref{tab:ablation}, removing either knowledge layer leads to a
performance drop.
These results confirm that factual knowledge provides evidence base for reasoning, while regularity knowledge facilitates mechanistic generalization.
Notably, the retrieved cases play a more critical role, as drug associations from KGs do not directly determine interaction types. 
Accurate prediction demands deeper insights into pharmacological mechanisms derived from historical cases, highlighting the importance of case-based reasoning.

\subsubsection{Effectiveness of Hybrid Case Retriever}
We evaluate the effectiveness of the hybrid retriever by varying the similarity weight $\lambda$ between semantic and structural components in \eqref{eq:retrieve}. Specifically, we measure the retrieval accuracy by selecting the top-$K$ cases ($K=1,5$) under different $\lambda$ values and assigning the majority label among them to the test sample. 
As shown in Figure~\ref{fig:hybrid}, retrieval accuracy first increases and then decreases as $\lambda$ changes, suggesting that a balanced combination of semantic and structural similarity yields optimal performance. This demonstrates that our hybrid retriever effectively integrates both drug functional descriptions and structural associations, enabling the retrieval of cases that are not only pharmacologically similar but also share interaction patterns, thereby improving the accuracy of predictions.

\subsubsection{Influence of Representative Sampling}

Table~\ref{tab:cluster} demonstrates the impact of our representative sampling strategy for case refinement. 
By replacing individual cases with representative cluster centroids, we significantly reduce the size of the case repository—by over 90\% in DrugBank—thus greatly enhancing scalability.
Notably, reducing the case volume does not compromise performance, while still achieving comparable or even improved results.
These results indicate the representative sampling strategy optimizes system efficiency and computational resource usage while filtering out noisy or redundant cases, leading to more representative and informative case selection. 

\subsection{Case Study}
\label{ssec:case}

\begin{figure}[t]
	\centering
	\vspace{-8px}
	\includegraphics[width=0.49\textwidth]{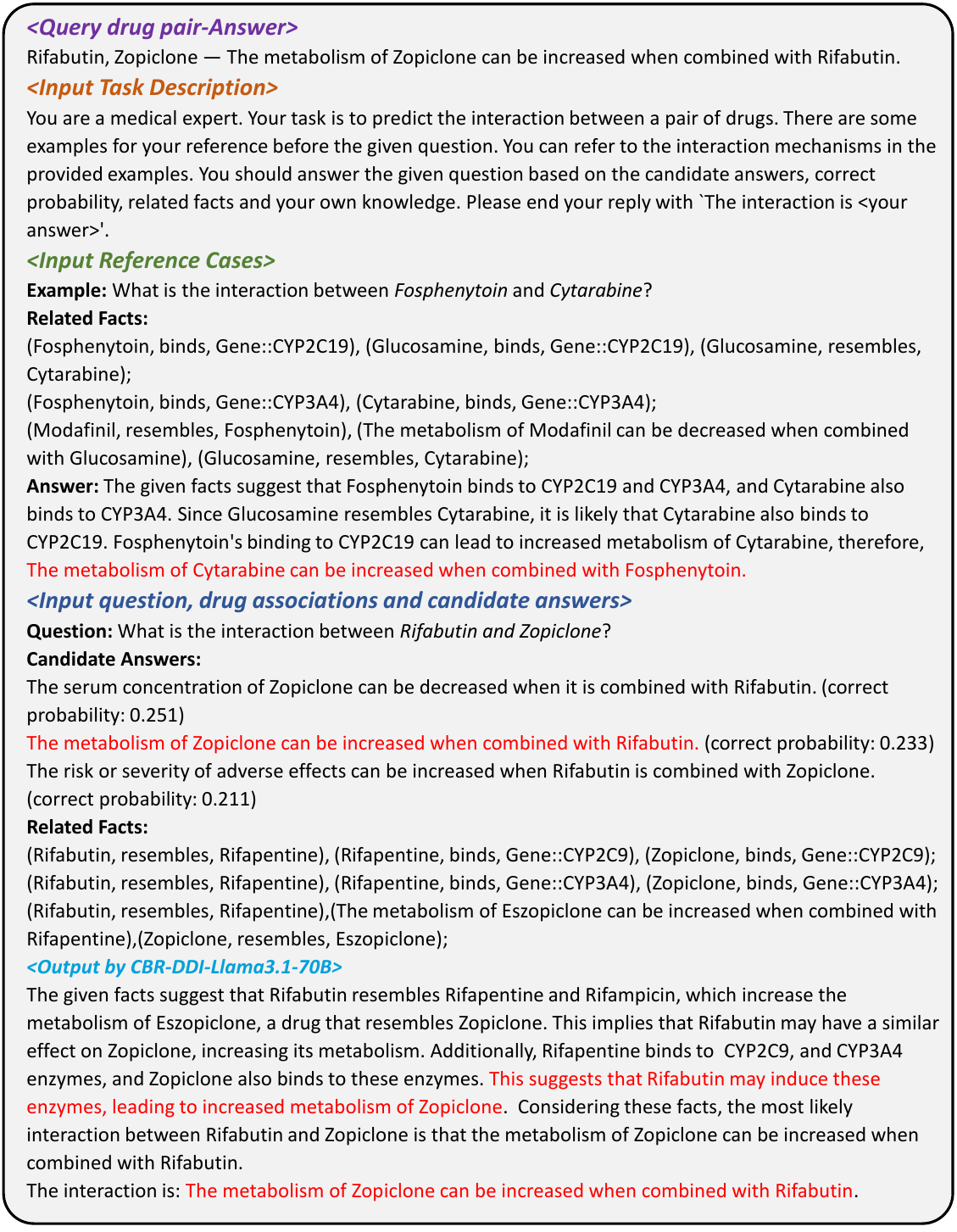}
	\vspace{-18px}
	\caption{One case study from DrugBank.}
	\label{fig:case}
	\vspace{-16px}
\end{figure}

We present a case study in Figure~\ref{fig:case}, which shows the query drug pair, input task description, one of the retrieved cases, extracted drug associations, filtered candidate answers, and the final output of the LLM.
As shown, the retrieved case exhibits similar drug associations and interaction mechanisms to those of the query pair, providing strong reasoning evidence.
The LLM leverages its powerful in-context learning capabilities to analyze the provided knowledge, generating accurate predictions and explanations, which provides useful insights for medical practitioners.
This example illustrates how CBR-DDI effectively enhances the LLM’s reasoning by incorporating valuable pharmacological knowledge from historical cases and KGs, resulting in accurate and faithful outcomes.

\section{Conclusion}

In this work, we introduce CBR-DDI, a novel framework that leverages CBR to enhance LLMs for DDI tasks. CBR-DDI constructs a knowledge repository by distilling pharmacological principles by LLM from historical cases and integrating structured knowledge extracted by GNN from KGs.
The framework employs comprehensive case retrieval, knowledge-enhanced case reuse, and dynamic case refinement, achieving accurate predictions, while maintaining high interpretability and flexibility.

\section*{Limitations}

%Since December 2023, a "Limitations" section has been required for all papers submitted to ACL Rolling Review (ARR). This section should be placed at the end of the paper, before the references. The "Limitations" section (along with, optionally, a section for ethical considerations) may be up to one page and will not count toward the final page limit. Note that these files may be used by venues that do not rely on ARR so it is recommended to verify the requirement of a "Limitations" section and other criteria with the venue in question.

In our approach, the prediction relies solely on textual information, without incorporating the drug molecular structures. This limits the model’s ability to perform fine-grained interaction analysis at a molecular level. 
In future work, it is worthy exploring how molecular structural data can be integrated into our framework, enabling more precise case retrieval and offering deeper pharmacological explanations of interaction mechanism.

%\section*{Acknowledgments}


%\bibliographystyle{plain}
\bibliography{custom}

\clearpage
\appendix

\section{Implementation Details}

\subsection{Details of Knowledge Repository}
\label{appendix:detail of repo}
\textbf{Repository Initialization.}
To initialize the knowledge repository, we randomly sample a subset of instances from the training data and use them to construct the initial set of cases. 
For each selected drug pair, we provide the LLM (e.g., Llama3.1-8B-Instruct) with the correct interaction type and relevant drug association facts, prompting it to generate a clear and accurate explanation of the underlying mechanism.

\noindent
\textbf{Repository Update.}
Whenever the number of cases in the knowledge base exceeds the threshold, or when a certain number of new cases (e.g., 1000) are added, we execute our representative sampling case refinement method.
Specifically,
we apply the K-Medoids clustering algorithm~\cite{park2009simple} within each DDI category to group semantically similar cases, using the text embeddings of their interaction mechanisms $M_c$. 
The number of clusters is pre-specified based on the overall sample size (e.g., retaining 5\% of the cases or at least 10 cases per category). 
Within each cluster, only the medoid—the most central and representative case—is retained, while redundant or overly similar cases are removed. 
This approach not only reduces storage and computational overhead but also ensures that the retained cases reflect diverse pharmacological scenarios.

\subsection{Algorithms for GNN module.}
\label{appendix:alg}

Following~\cite{zhang2023emerging}, we present the algorithms of the GNN module.
Given a drug pair $p = (u,v)$,
we implicitly encode the pair-wise subgraph representations with Algorithm~\ref{alg:emergnn},
and use beam search to find the top-$P$ paths between them with Algorithm~\ref{alg:path}.

\begin{algorithm}[ht]
	\caption{Pair-wise subgraph representation learning with flow-based GNN.}
	\label{alg:emergnn}
	\small
	\begin{algorithmic}[1]
		\REQUIRE {$p=(u, v), \bm f_u=f(D_u), \bm f_v=f(D_v), L, \delta, \sigma, \{\bm W^{(\ell)}, \bm w^{(\ell)}\}_{\ell=1\dots L}\}$, $\mathcal{G}$.\\} \COMMENT{$p=(u, v)$: drug pair; $\{\bm f_u, \bm f_v\}$: the embeddings of drug descriptions; $L$: the  depth of path-based subgraph; $\delta$: activation function; $\sigma$: sigmoid function; $\{\bm W^{(\ell)}, \bm w^{(\ell)}\}_{\ell=1\dots L}\}$: learnable parameters; $\mathcal{G}$: biomedical KG.}
		\STATE initialize the $u\rightarrow v$ pair-wise representation as $\bm h_{u,e}^0 = \bm f_u$ if $e=u$, otherwise $\bm h_{u,e}^0 = \bm 0$;
		\STATE initialize the $v\rightarrow u$ pair-wise representation as $\bm h_{v,e}^0 = \bm f_v$ if $e=v$, otherwise $\bm h_{v,e}^0 = \bm 0$;
		\FOR{$\ell\leftarrow 1$ to $L$}
		\FOR[This loop can work with matrix operations in parallel.]{$e\in\mathcal V_\text{D}$}
		\STATE message for $u\rightarrow v$: \\
		{\small{$\begin{aligned}
			\bm h_{u,e}^{(\ell)} = \delta\biggl(\bm W^{(\ell)}\sum_{(e',r,e)\in\mathcal N_\text{D}} \!
			\sigma\left((\bm w_r^{(\ell)})^\top [\bm f_u; \bm f_v]\right)\cdot \\
			 \left(\bm h_{u,e'}^{(\ell-1)}\odot\bm h_{r}^{(\ell)} \right) \biggr)
		\end{aligned}$;
		\STATE message for $v\rightarrow u$: \\
		$\begin{aligned}
			\bm h_{v,e}^{(\ell)} = \delta\biggl(\bm W^{(\ell)}\sum_{(e',r,e)\in\mathcal N_\text{D}} \! 
			\sigma\left((\bm w_r^{(\ell)})^\top [\bm f_u; \bm f_v]\right)\cdot \\
			 \left(\bm h_{v,e'}^{(\ell-1)}\odot\bm h_{r}^{(\ell)} \right) \biggr)
		\end{aligned}$;}}
		\ENDFOR
		\ENDFOR
		\STATE \textbf{Return} $\bm h_p = [\bm h_{u,v}^{(L)}; \bm h_{v,u}^{(L)}]$.
	\end{algorithmic}
\end{algorithm}

\begin{algorithm}[ht]
	\caption{Path extractor.}
	\label{alg:path}
	\small
	\begin{algorithmic}[1]
		\REQUIRE {$(u,v), L, P$}
		\STATE initialize openList$[0]\leftarrow u$;
		\STATE set $\mathcal V_{u,v}^{(0)}=\{u\}, \mathcal V_{u,v}^{(L)}=\{v\}$;
		\STATE obtain the set $\mathcal V_{u,v}^{(\ell)}=\{e: d(e,u)=\ell, d(e,v)=L-\ell\}, \ell=1, \dots, L$ with bread-first-search;
		\FOR{$\ell\leftarrow1$ to $L$}
		\STATE set closeList$[\ell]\leftarrow \emptyset$, pathList$[\ell]\leftarrow \emptyset$;
		\FOR{each edge in $\{(e',r,e): e'\in\text{openList}[\ell-1], e\in\mathcal V_{u,v}^{\ell} \}$}
		\STATE compute the attention weights $\alpha_{r}^{(\ell)} = \sigma\left((\bm w_{r}^{(\ell)})^\top[\bm f_{u};\bm f_{v}]\right)$;
		\STATE compute score($u,e', e$) = score($u,e$) + $\alpha_{r}^{(\ell)}$;
		\STATE closeList[$\ell$].add(($e$, score($u,e',e$)));
		\ENDFOR
		\FOR{$(u,e',e)\in$top$_P$(clostList[$\ell$])}
		\STATE openList[$\ell$].add($e$),  pathList[$\ell$].add($(e',r,e)$);
		\ENDFOR
		\ENDFOR
		\STATE \textbf{Return:} join(pathList[$1$]$\dots$pathList[$L$]).
	\end{algorithmic}
\end{algorithm}

\subsection{Details of Experiments}
\label{appendx:exp}

\textbf{Datasets.}
We conduct experiments on two widely used DDI datasets: (1) DrugBank~\cite{drugbankwishart2018drugbank}, a multiclass DDI prediction dataset that contains 86 types of pharmacological interactions between drugs. 
(2) TWOSIDES~\cite{twotatonetti2012data}, 
a multilabel DDI prediction dataset that records 200 side effects between drugs.
We use HetioNet~\cite{kg1himmelstein2015heterogeneous} as for the external biomedical knowledge graph.
Table~\ref{tab:data} and \ref{tab:kg}
display the statistics of the datasets and knowledge graph, where $\mathcal V$'s represent the sets of nodes, $\mathcal R$’s represent the sets of interaction types, and $\mathcal N$’s represent the sets of edges.

\begin{table*}[ht]
	\setlength{\tabcolsep}{3pt}
	\centering
	\small
	\begin{tabular}{lccccccccc}
		\toprule
		\multicolumn{1}{l}{\multirow{2}{*}{Dataset}} & \multirow{2}{*}{$|\mathcal{V}_{\text{D-train}}|$} & \multirow{2}{*}{$|\mathcal{V}_{\text{D-valid}}|$} & \multirow{2}{*}{$|\mathcal{V}_{\text{D-test}}|$} & \multirow{2}{*}{$|\mathcal{R}_{\text{D}}|$} & \multirow{2}{*}{$|\mathcal{N}_{\text{D-train}}|$} & \multicolumn{2}{c}{S1}                 & \multicolumn{2}{c}{S2}                 \\
		\multicolumn{1}{c}{}     &       &     &       &      &          & \multicolumn{1}{c}{$|\mathcal{N}_{\text{D-valid}}|$} & $|\mathcal{N}_{\text{D-test}}|$ & \multicolumn{1}{c}{$|\mathcal{N}_{\text{D-valid}}|$} & $|\mathcal{N}_{\text{D-test}}|$ \\
		\midrule
		DrugBank     & 1,461                      & 79                          & 161                       & 86                  & 137,864                    & {17,591}   & 32,322  &{536}      & 1,901   \\ 
		TWOSIDES                                    & 514                        & 30                          & 60                        & 200                 & 185,673                    & {3,570}    & 6,698   & {106}      & 355     \\ 
		\bottomrule
	\end{tabular}
	\vspace{-4px}
	\caption{Statistics of datasets.}
		\label{tab:data}
	\vspace{-8px}
\end{table*}

\begin{table}[ht]
	\centering
	\begin{tabular}{lccc}
		\toprule
		KG & $|\mathcal{V}_{\text{B}}|$ & $|\mathcal{R}_{\text{B}}|$ & $|\mathcal{N}_{\text{B}}|$ \\
		\midrule
		HetioNet & 34,124 & 23 & 1,690,693 \\
		\bottomrule
	\end{tabular}
	\vspace{-4px}
	\caption{Statistics for knowledge graph.}
	\label{tab:kg}
	\vspace{-8px}
\end{table}

\noindent
\textbf{Evaluation metrics.}
For the DrugBank dataset,
there is one interaction between a pair of drugs.
Hence, we evaluate the performance in a multi-class setting,
which estimates whether the model can correctly predict the interaction type for a pair of drugs.
We consider the following metrics:
\begin{itemize}[leftmargin=*]
	\item Accuracy: the percentage of correctly predicted interaction type compared with the ground-truth interaction type.
	\item $\text{F1(macro)}=\frac{1}{\|\mathcal I_D\|}\sum_{i\in\mathcal I_D}\frac{2P_i\cdot R_i}{P_i+R_i}$,
	where $P_i$ and $R_i$ are the precision and recall for the interaction type $i$, respectively.
	The macro F1 aggregates the fractions over different interaction types.
\end{itemize}

In the TWOSIDES dataset,
there may be multiple interactions between a pair of drugs, such as anaemia, nausea and pain.
Hence, we treat it as a recommendation task, where the LLM is prompted to recommend 5 possible interactions for given drug pair. We use Recall@5 and NDCG@5 as the evaluation metrics:
\begin{align}
	\text{Recall}@5 &= \frac{{|R_{1:5}\cap T|}}{|T|},	 
	\\
	\text{NDCG}@5 &= \frac{\sum_{i=1}^{5}\mathbb{I}(R_i\in T)\nicefrac{1}{\log_2 (i+1)}}{\sum_{i=1}^{\min(|T|,5)}\nicefrac{1}{\log_2 (i+1)}},
\end{align}
where $R$ is a list of recommended
interactions for the given pair, $T$ is the ground-truth list, and indicator function $I(x) = 1$ if $x$ is true and $0$ otherwise.

\noindent
\textbf{Hyperparameters.}
For the training of the GNN module, we follow EmerGNN~\cite{zhang2023emerging}'s hyperparameter settings.
We use three LLMs in experiments: Llama3.1-8B-Instruct~\cite{grattafiori2024llama}, Llama3.1-70B-Instruct~\cite{grattafiori2024llama},  
and DeepSeek-V3~\cite{liu2024deepseek}.
The training of GNN module and the inference of Llama3.1-8B are on an RTX 3090-24GB GPU, while the inference for Llama3.1-70B runs on two A100-80GB GPUs. 
DeepSeek is accessed via API calls.
We set the number of reference cases $K$ to 5, maintain $P=5$ paths in drug associations, and limit candidate answers to 3 for DrugBank and 10 for TWOSIDES.

\noindent
\textbf{Baseline Methods.}
We consider following baseline methods for performance comparison: 

(1) traditional methods without using LLMs:
\begin{itemize}
	\item MLP~\cite{mlp1998artificial} uses multilayer perceptron to map the fingerprint features of drugs to the interaction types between them.
	\item ComplEx~\cite{complextrouillon2017knowledge} converts KG in to a complex matrix and predict DDI based on the decomposition of the matrix.
	\item MSTE~\cite{yao2022effective} is an embedding-based method that learns on KG to predict the possibility of whether a relation exists.
	\item Decagon~\cite{Decagonzitnik2018modeling} utilizes drug, genes and diseases information to learn drug representation and predict DDI with a graph convolutional network.
	\item SumGNN~\cite{sumgnn2021} samples a subgraph from KG for drug pair and designs a summarization scheme to generate reasoning path in the subgraph. 
	\item EmerGNN~\cite{zhang2023emerging} designs a flow-based GNN on the KG to learn the representation of subgraph between drugs for prediction.
	\item TIGER~\cite{tigersu2024dual} uses graph transformer to encode the molecular structure and biomedical KG to learn dual-channel representation for drugs.	
	\item TextDDI~\cite{textddizhu2024learning} trains an LM as predictor with an RL-based information selector for extracting relevant drug descriptions.
\end{itemize} 

(2) LLM-based methods:
\begin{itemize}
	\item Base model is a zero-shot method which directly prompts LLMs to select the most likely interaction type $r$ from the relation set $\mathcal{R_D}$.
	\item Naive-CBR~\cite{brown2020language} retrieves 10 similar labeled cases based on fingerprint similarity as few-shot prompting.
	\item K-Paths~\cite{kpath} employs a diversity-aware adaptation of Yen’s algorithm to retrieve
	the K shortest paths between drugs for LLM's prediction.
\end{itemize}

\section{Supplementary Experiments}

\subsection{Performance on S0 Setting}
\label{appendix:S0}
\begin{table}[t]
	\centering
	%	\vspace{-8px}
	%	\renewcommand\arraystretch{0.98}
	\setlength{\tabcolsep}{5pt}
	\footnotesize
		\scalebox{0.9}{
		\begin{tabular}{c|l|cc|cc}
			\toprule
			\multirow{2}{*}{Type} & \multirow{2}{*}{Method} & \multicolumn{2}{c|} {DrugBank} & \multicolumn{2}{c}{ {TWOSIDES}}\\                                                 
			& & {Acc}   & {F1}    & {Recall}    & {NDCG}     \\
			\midrule
			Feature-based &MLP & 81.22 &  61.56 & 25.21 & 27.78 \\
			\midrule
			\multirow{3}{*}{Graph-based}
			&Decagon & 87.10 & 58.61 & 12.47 & 14.92\\
			&{EmerGNN} & {96.48} & 95.44 & {26.84} & 30.22   \\
			&TIGER   & 95.57 & 93.89 & 21.54 & 25.36\\
			\midrule
			LM-based&TextDDI & 96.04 & 94.53  & 14.07 & 17.64 \\
			\midrule
			\multirow{4}{*}{Llama3.1-70B}& Base  & 9.17 & 4.79 & 0.06 & 0.07 \\
			&  Naive-CBR  & 57.92 & 54.26 & 7.05 & 8.74 \\
			& K-Paths  & 23.75 & 15.27 & 0.87 & 1.38 \\
			& CBR-DDI  & \textbf{96.98} & \textbf{95.95} & \textbf{27.18} &\textbf{31.04} \\
			\bottomrule
		\end{tabular}
	}
	\vspace{-6px}
	\caption{\label{tab:result-S0}
		Performance comparison of different methods for DDI on S0 setting.}
	\vspace{-8px}
\end{table}

We present the performance of different methods under the S0 setting (predicting interactions between existing drugs) in Table~\ref{tab:result-S0}. 
As can be seen, our method still achieves the best performance. 
However, the advantage is not as pronounced as in the S1 and S2 settings, since our approach primarily targets the scenario of new drug prediction. 
Under the S0 setting, existing methods can memorize possible interaction types between known drugs through training, whereas our method does not fine-tune LLMs and thus lacks this advantage.

\subsection{Effect of Case Number}

\begin{figure}[t]
	\centering
	\includegraphics[width=0.47\textwidth]{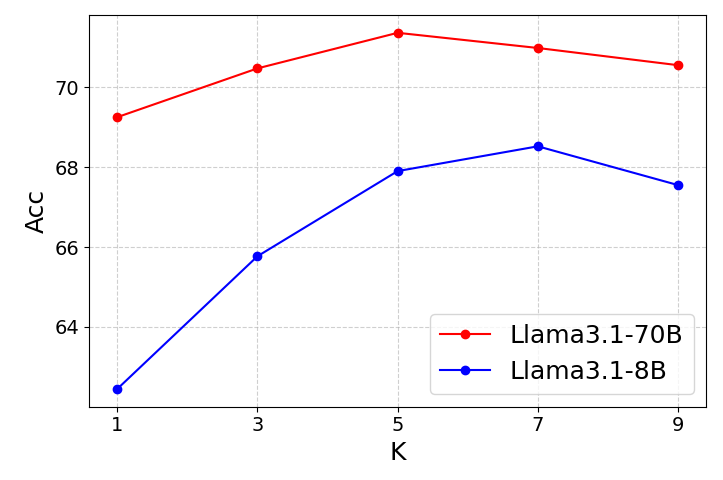}
	\vspace{-5px}
	\caption{Impact of the number of retrieved cases on DrugBank-S1.}
	\label{fig:K}
	\vspace{-5px}
\end{figure}

We investigate how the number of retrieved cases $K$ affects model performance. 
As shown in Figure~\ref{fig:K}, increasing $K$ generally improves accuracy for both the Llama3.1-8B and 70B models. These results suggest that incorporating more cases enhances LLM's reasoning by providing richer phamacological insights, but overly large $K$ may introduce redundancy or noise. 
Specifically, incorporating case information can significantly enhance the performance of smaller LLMs (i.e., Llama3.1-8B), as their weaker reasoning capabilities make it difficult to delve beyond superficial drug associations to uncover underlying interaction mechanisms and consequently make accurate predictions.

\subsection{Effect of Drug Association Knowledge}

\begin{figure}[t]
	\centering
	\includegraphics[width=0.47\textwidth]{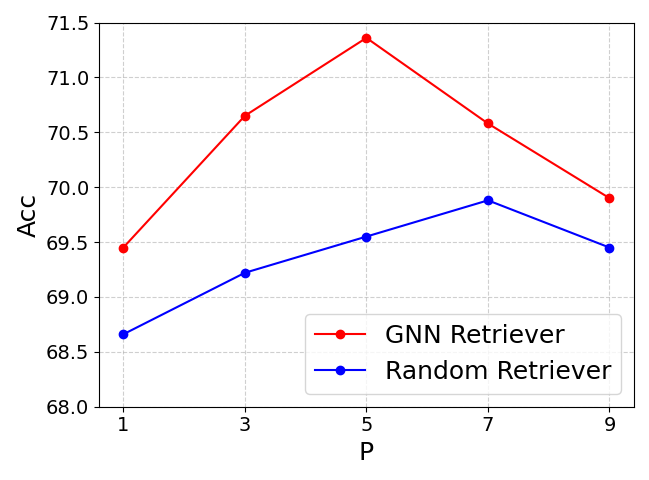}
	\vspace{-5px}
	\caption{Impact of retrieved drug associations on DrugBank-S1 of CBR-DDI-Llama3.1-70B.}
	\label{fig:P}
	\vspace{-5px}
\end{figure}

We also analyze the impact of the number of extracted drug association paths $P$ on model performance. 
As shown in Figure~\ref{fig:P}, prediction accuracy initially improves with increasing $P$, as additional paths provide more factual evidence for mechanistic reasoning. 
However, beyond an optimal point, performance gradually declines as excessive paths introduce irrelevant or conflicting relationships that obscure core interaction mechanisms. 

Furthermore, the Figure compares our attention-based GNN retriever with the random retriever (i.e., heuristic retrieval used in existing methods). 
The results demonstrate that our GNN retriever achieves superior performance, as the attention mechanism enables the model to learn and prioritize more high-quality relationship paths, thereby providing a more effective foundation for reasoning. 
In contrast, heuristic retrieval methods lack this discriminative capability to identify the critical pharmacological relationships.

\subsection{Effect of Hybrid Retriever}

\begin{figure*}[ht]
	\centering
	\includegraphics[width=0.80\textwidth]{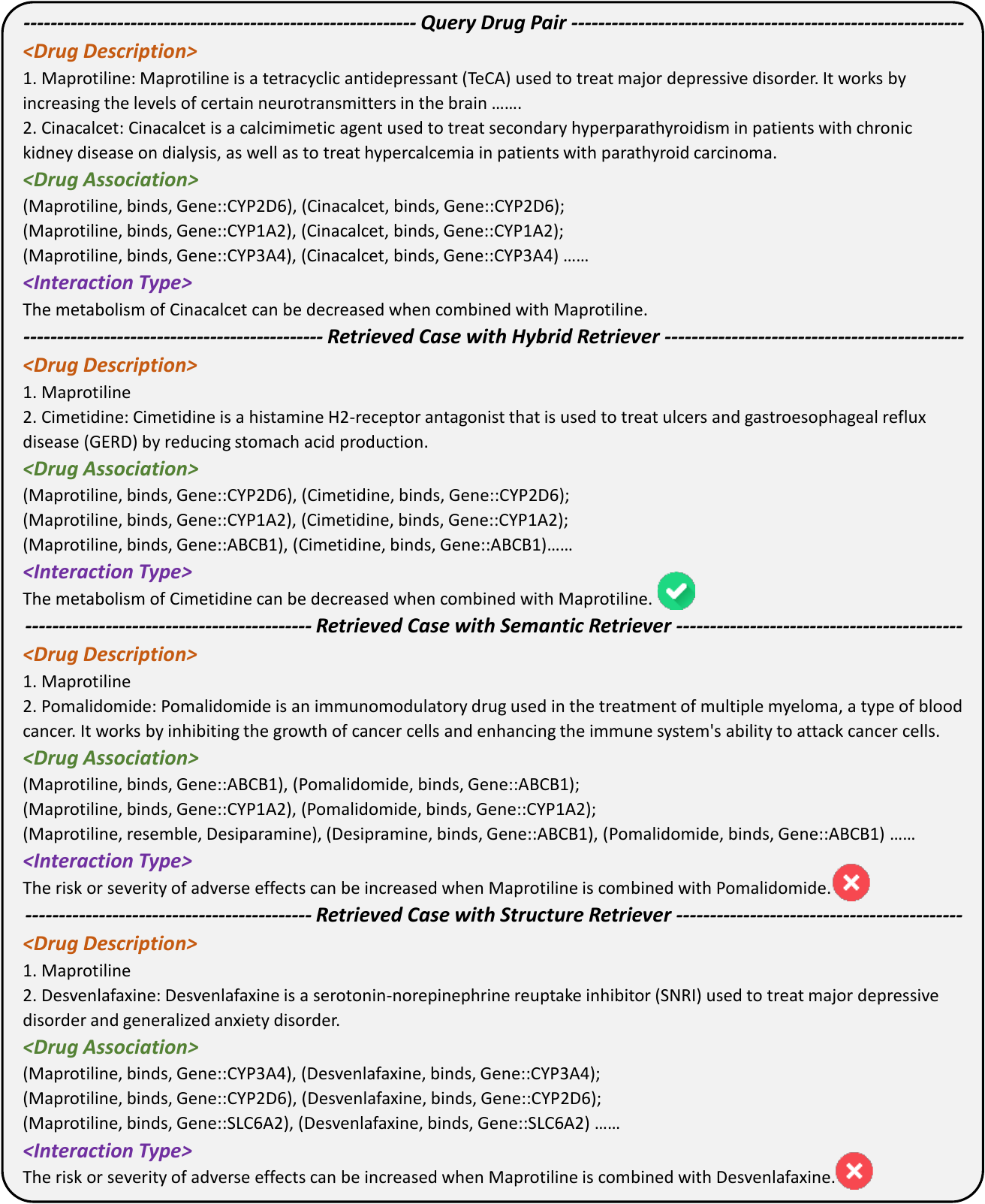}
	\vspace{-5px}
	\caption{Retrieved cases of different retrievers on DrugBank-S1.}
	\label{fig:retrieved case}
	\vspace{-5px}
\end{figure*}

We present the most relevant cases retrieved by different retrievers for the same query drug pair. As shown in Figure~\ref{fig:retrieved case}, using either the semantic-based retriever ($\lambda=1$) or the structure-based retriever ($\lambda=0$) alone fails to effectively retrieve cases with the same interaction type as the test case, thus unable to provide valuable interaction mechanisms for the LLM. 
In contrast, our proposed hybrid retriever combines semantic similarity and structural similarity, capturing relevant pharmacological effects and drug associations to deliver meaningful pharmacological insights. 
Note that we do not display the interaction mechanisms in the cases here, as they are not involved in the retrieval process.

\section{Case Study}
\label{appendix:case}

\begin{figure*}[ht]
	\centering
	\includegraphics[width=0.80\textwidth]{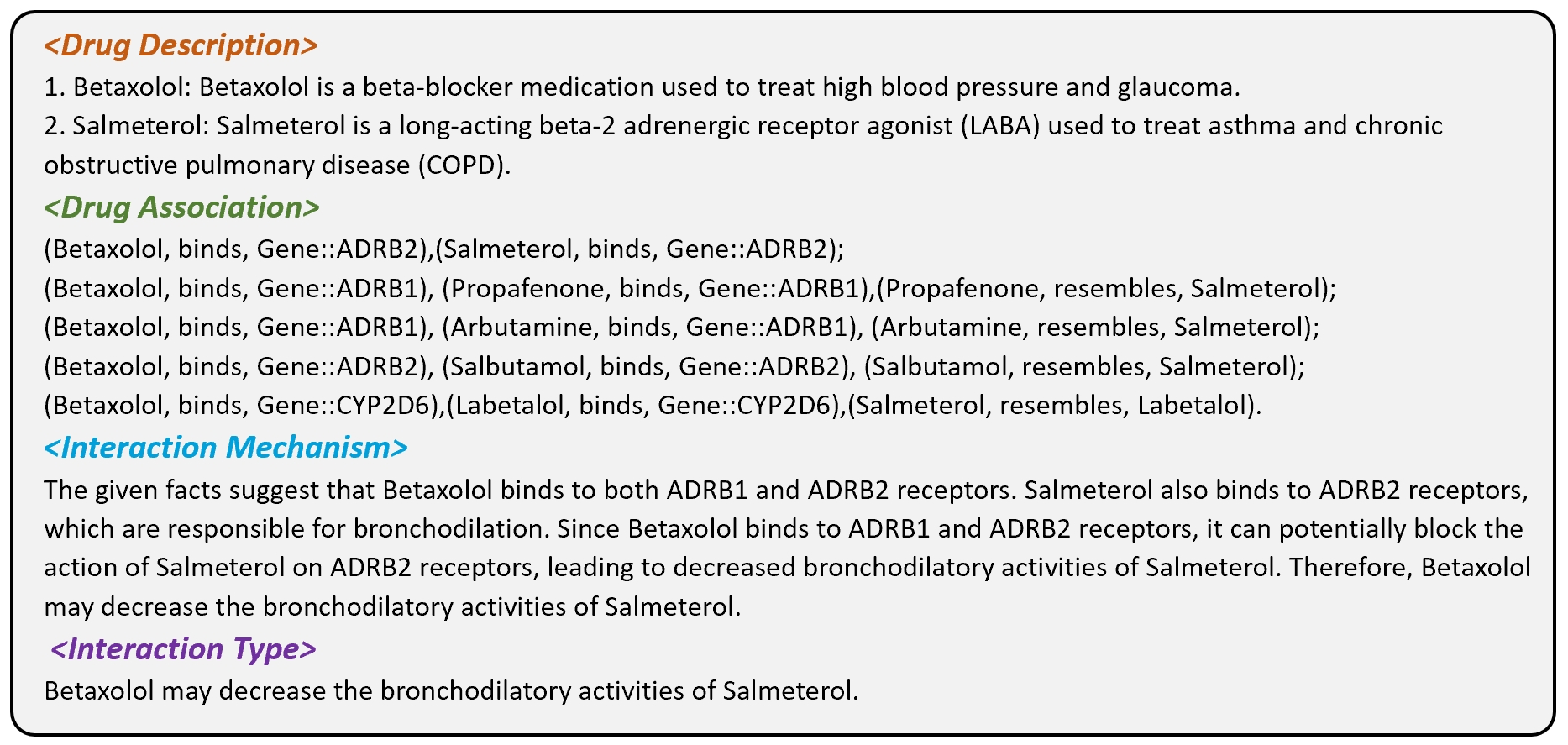}
	\vspace{-5px}
	\caption{One case from DrugBank.}
	\label{fig:case1}
	\vspace{-5px}
\end{figure*}

\begin{figure*}[ht]
	\centering
	\includegraphics[width=0.80\textwidth]{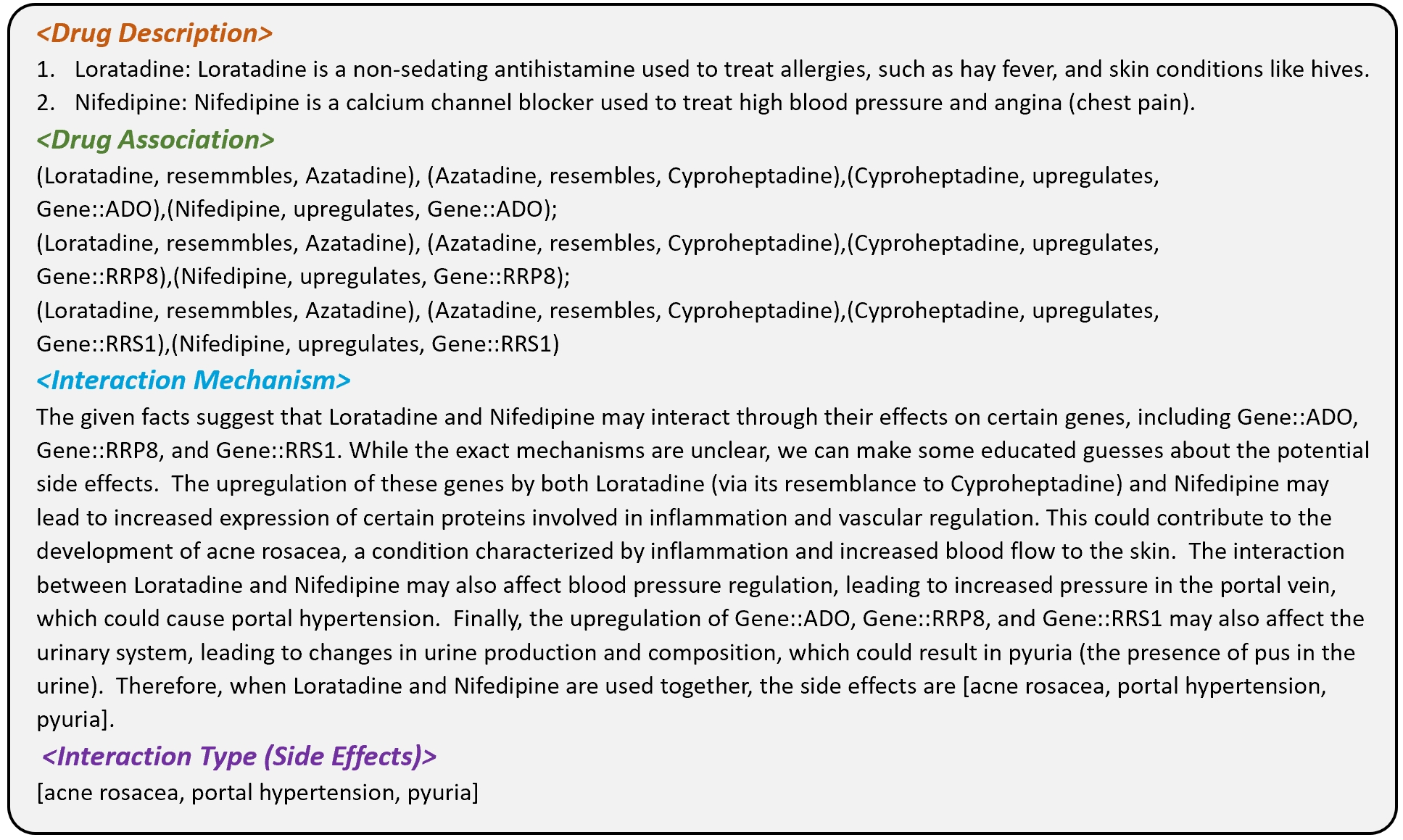}
	\vspace{-5px}
	\caption{One case from TWOSIDES.}
	\label{fig:case2}
	\vspace{-5px}
\end{figure*}

We present two more representative cases from DrugBank and TWOSIDES in Figure~\ref{fig:case1} and Figure~\ref{fig:case2}. 
Each case includes LLM-generated drug descriptions, key drug associations extracted by the GNN module, mechanistic explanations generated by the LLM based on both external and internal knowledge, and the ground truth interaction label. 
These cases are constructed to capture both factual evidence and underlying pharmacological principles of drug interactions, thereby supporting accurate retrieval and interpretable reasoning for new prediction tasks.

\end{document}